\documentclass[preprint,12pt]{elsarticle}
\newtheorem{remark}{Remark}
\biboptions{sort&compress}
\usepackage{amssymb}
\usepackage{amsmath}
\usepackage{algorithm}
\usepackage{algpseudocode}
\usepackage{multirow}
\usepackage{subfigure,subcaption}
\usepackage{amsfonts,amssymb,color}

\journal{Engineering Applications of Artificial Intelligence}

\begin{document}

\begin{frontmatter}

\title{Intrinsic-Motivation Multi-Robot Social Formation Navigation with Coordinated Exploration}

\author[label1,label2]{Hao~Fu\corref{cor1}}
        \author[label1,label2]{Wei~Liu}
        \author[label1,label2]{Shuai~Zhou}
        \address[label1]{School of Computer Science and Technology, Wuhan University of Science and Technology, Wuhan 430081, China}
        \address[label2]{Hubei Province Key Laboratory of Intelligent Information Processing and Real-time Industrial System, Wuhan 430081, China}
        \tnotetext[t1]{This work was supported in part by the National Natural Science Foundation of China under Grant 62303357 and Grant 62173262 and in part by the Hubei Provincial Natural Science Foundation of China under Grant 2023AFB109.}
        \cortext[cor1]{Corresponding author: Hao Fu, Email: fuhao@wust.edu.cn.}

\begin{abstract}
This paper investigates the application of reinforcement learning (RL) to multi-robot social formation navigation, a critical capability for enabling seamless human-robot coexistence. While RL offers a promising paradigm, the inherent unpredictability and often uncooperative dynamics of pedestrian behavior pose substantial challenges, particularly concerning the efficiency of coordinated exploration among robots. To address this, we propose a novel coordinated-exploration multi-robot RL algorithm introducing an intrinsic motivation exploration. Its core component is a self-learning intrinsic reward mechanism designed to collectively alleviate policy conservatism. Moreover, this algorithm incorporates a dual-sampling mode within the centralized training and decentralized execution framework to enhance the representation of both the navigation policy and the intrinsic reward, leveraging a two-time-scale update rule to decouple parameter updates. Empirical results on social formation navigation benchmarks demonstrate the proposed algorithm's superior performance over existing state-of-the-art methods across crucial metrics.  Our code and video demos are available at: https://github.com/czxhunzi/CEMRRL.
\end{abstract}

%% Keywords

\end{frontmatter}

\section{Introduction}
\label{sec1}
%% Labels are used to cross-reference an item using \ref command.

As robotics and artificial intelligence make major advances, autonomous robot navigation has garnered considerable attention, such as smart production \cite{jiang2024geometry}, logistics optimization \cite{lee2022research}, and autonomous vehicle systems \cite{lin2025velocity}. While single-robot systems have demonstrated significant progress, their inherent limitations in scalability, flexibility, and robustness have motivated the exploration of multi-robot systems for more complex tasks. Within this landscape, multi-robot social cooperative navigation has emerged as a particularly promising paradigm for tackling complex tasks in human-populated environments \cite{DongL2024}. This paradigm leverages the collective capabilities of multiple robots to achieve goals that would be challenging or impossible for a single agent, with transformative potential in autonomous driving~\cite{ToghiB2022}, object search~\cite{DalmassoM2021}, and formation control~\cite{SuiZ2021,FuH2024}. Social formation navigation, the focal point of this study, remains constrained by two key challenges: (1) limited human-robot communication and (2) unpredictable pedestrian dynamics in human-populated environments \cite{TalebpourZ2018}. These factors complicate the design of robust social cooperative navigation algorithms.

In response to these challenges, the traditional methods for social cooperative navigation predominantly rely on the reaction-based methods, such as reciprocal velocity obstacle (RVO) \cite{BergJur2008}, social force model (SFM) \cite{FerrerG2013}, optimal reciprocal collision avoidance (ORCA) \cite{VanDenB2011}. However, these methods are limited in their ability to interpret interaction in social navigation scenarios, often leading to shortsighted and occasionally unnatural behaviors. In contrast, trajectory-based methods \cite{NavsalkarA2023,LeV2024} attempt to improve foresight by predicting pedestrian trajectories to plan multi-robot paths. Unfortunately, these prediction models are computationally expensive, hindering real-time applications. Without explicitly online prediction, imitation learning approaches have been developed, such as \cite{ChenL2023}. It should be mentioned that, those methods exhibit poor adaptation to previously unseen environments, whereas our proposed algorithm bridges this gap by introducing a multi-robot coordinated exploration  strategy that enables adaptive exploration in the social formation navigation scenarios.

As an alternative, recent developments in deep reinforcement learning (DRL) have provided a novel avenue for improving adaptation to prior unseen scenarios through continual interactions between human and machine~\cite{TanJ2025b,ChenY2017,ChenY2017b} or among machines~\cite{TanJ2025,WangW2024}. Owing to this inherent capability, DRL has emerged as a promising solution for the social cooperative navigation and other intelligent decision-making tasks, including power internet of things~\cite{ZhangS2022}, transportation~\cite{QiaoF2021}, energy management~\cite{ZhangH2022,ZhangH2024}. In particular, for social cooperative navigation, Chen et al. \cite{ChenY2017} pioneered a collision avoidance with DRL (CADRL), demonstrating adaptability to previously unseen social navigation environments. Subsequent extensions incorporated social norms to develop a socially-aware CADRL framework \cite{ChenY2017b}. Further advancements include CADRL-based formation control \cite{SuiZ2021}, though these approaches still fall short of fully capturing the intricacies of social interactions. To address this limitation, recent studies \cite{DongL2024,ZhangT2022,WangW2024} have explored the application of attention mechanisms to dynamically assign importance to social relationships by combining with multi-agent reinforcement learning (MARL). Moreover, to eliminate inter-robot communication dependencies inherent resulting from the centralized manner in \cite{SuiZ2021}, training multiple robots to reach their team objectives with the centralized training and decentralized execution (CTDE) paradigm~\cite{RashidT2018,LoweR2017} achieves better scalability. Additionally, \cite{SongC2024} also designed a local-and-global attention module to extract intra-agent and inter-agent critical environmental information. 

It is noteworthy that the aforementioned works primarily rely on the $\epsilon$-greedy exploration strategy. Considering the inherent unpredictability and potentially uncooperative nature of pedestrian movements, such an exploration strategy often results in inefficient and redundant stochastic exploration. Beyond the stochastic exploration, significant contributions have been made by evolutionary algorithms (EAs) in addressing exploration challenges. EAs employ principles of mutation and crossover to evolve populations of candidate solutions. For instance, EAs have been successfully applied to co-optimize component sizing and energy management for hybrid powertrains \cite{LeiN2023}. Furthermore, more recent advancements in this field demonstrate their efficacy in diverse applications, including multi-robot task assignment \cite{DongJ2024}, ammonia-hydrogen propulsion systems \cite{ZhangH2024b}, and path planning \cite{HengH2025}.
While EAs are powerful for exploration, their direct application to the online real-time multi-robot social navigation scenarios with continuous state and action spaces can be computationally intensive. The employment of soft actor-critic (SAC) in \cite{HeZ2023} offers a potential mitigation to this limitation by obviating the need for evaluating an entire population and bypassing stochastic exploration. However, its independent exploration behavior generated by SAC suffers from the ill-coordinated exploration. This shortfall exacerbates the problem of relative overgeneralization in multi-robot social cooperative navigation. 

To address the identified gap, we propose a coordinated-exploration multi-robot reinforcement learning (CEMRRL) algorithm, integrating an intrinsic motivation exploration criterion within the social formation navigation---a specific category of multi-robot social cooperative navigation problems. In summary, the key contributions of this work are summarized as follows.

(1) This paper establishes a self-learning intrinsic reward mechanism by combining self-adjusting joint policy entropy with an exploration bonus, improving multi-robot exploration. Adding a novelty differential function into the bonus in our established mechanism helps multiple robots coordinate better, reducing overly conservative navigation behaviors, compared with \cite{SongC2024,HeZ2023}.

(2)  A two-time-scale update rule is leveraged to decouple interdependent updates for the intrinsic reward and actor-critic parameters. Furthermore, the incorporation of a dual-sampling mode into the CTDE framework ensures the implementation of this process.

(3) Experimental results on several challenging benchmarks demonstrate that our algorithm outperforms previous state-of-the-art methods and presents markedly superior performance in the social formation navigation.

\section{Methodology}
\label{sec3}

In this section, the multi-robot social formation navigation problem is described. Then, we propose a CEMRRL algorithm based on the CTDE framework.

%% Use \subsection commands to start a subsection.
\subsection{Problem formulation}
\label{subsec1}
The multi-robot social formation navigation task can be treated as a sequential decision-making problem, which is described as
\begin{equation} \label{optim}
    \left\{
    \begin{aligned}
        & {\rm argmin}_{\pi} \ \mathbb{E} [t_{g}, \sum_{0}^{t_{g}}\sum_{i=1}^{n} \Vert p^{i} - p^{0} - p_{f}^{i} \Vert], \\ 
        & \Vert p^{i} - p^{-i} \Vert \ge d^{i} + d^{-i},  \   \forall   i=0,1,...,n,  \\ 
        & [p_{x}^{i},p_{y}^{i},\psi^{i}]|_{t} = [p_{x}^{i},p_{y}^{i},\psi^{i}]|_{t-\Delta t} + {\Delta t}  [\cos(\psi^{i})v^{i},\sin(\psi^{i})v^{i}, w^{i}]|_{t-\Delta t}, \\ 
        &  p^{0}(t_{g}) =  \bar{p}, \\ 
    \end{aligned}
    \right.
\end{equation}
where $i=0$ refers to the leader and $n$ to the number of followers, $\pi$, $p^{i}=[p_{x}^{i},p_{y}^{i}]$, $\bar{p}=[\bar{p}_{x},\bar{p}_{y}]$, $\psi^{i}$, $p_{f}^{i}$, $d^{i}$, $\Delta t$, $v^{i}$, and $w^{i}$ denote the joint policy, the position, the goal position, the heading angle, the relative position vector with respect to the leader, the radius, the sampling time interval, the line velocity, and the angle velocity for robot $i$, respectively. $-i$ denotes all agents except agent $i$. 

The agent’s state vector $s^{i}$ can be divided into an observable component  $s^{i}_{o}$ and an unobservable (hidden) one $s^{i}_{h}$. For the follower $i=1,2,...,n$ or the pedestrian $i=n+1,n+2,...$, $s^{i}_{o}=[p_{x}^{i},p_{y}^{i},v_{x}^{i},v_{y}^{i}, r^{i}] \in \mathcal{R}^{5}$ and $s^{i}_{h}=[\bar{v}^{i}, \psi^{i}] \in \mathcal{R}^{2}$, where $v_{x}^{i}$, $v_{y}^{i}$, and $\bar{v}^{i}$ are the velocity in the x-axis direction, that one in the y-axis direction, and the preferred speed, respectively. For the leader, $s^{0}_{o}=[p_{x}^{0},p_{y}^{0},v_{x}^{0},v_{y}^{0}, r^{0}] \in \mathcal{R}^{5}$ and $s^{0}_{h}=[\bar{p}_{x},\bar{p}_{y},\bar{v}^{0}, \psi^{0}] \in \mathcal{R}^{4}$.

After that, the sequential decision-making problem is transferred into a Decentralized Partially Observable Markov Decision Process (Dec-POMDP), which is defined by a tuple  $(\mathcal{I}, \mathcal{S}, \mathcal{O}, \mathcal{Z}, \mathcal{A}, \mathcal{P}, \mathcal{R}, \gamma)$, where $\mathcal{I}$ is the set of $n+m+1$ agents, $\mathcal{S}$ is the joint state space, $\mathcal{O}=\Pi_{i=0}^{n} \mathcal{O}_{i}$ is the joint observation space, $\mathcal{Z}$ is the joint observation probability, $\mathcal{A}=\Pi_{i=0}^{n} \mathcal{A}_{i}$ is the joint action space with the local action space $\mathcal{A}_{i}$, $\mathcal{P}$ is the state transition probability, $\mathcal{R}$ is the joint extrinsic reward, $\gamma \in (0,1)$ is the discount factor, and $\mathcal{O}_{i}$ denotes the partial observation space.

\subsubsection{Observation Space $\mathcal{O}$}
\label{subsec4}

The joint state $s \in \mathcal{S}$ is described by $s = [s^{0}, s^{1}, ...]$. For robot $i$, its local observation $o^{i} \in \mathcal{O}_{i}$ is represented by
\begin{align}
    o^{i} = [s^{i}, s^{-i}_{o}]. \ \forall i=0,1,...,n
\end{align}

When robots make decisions, a local action-observation history or trajectory ~\cite{FoersterJ2018} for robot $i$, $\tau^{i} \in \mathcal{T}  \equiv (\mathcal{O} \times \mathcal{A})^{*} $, is usually used to replace its local observation $o^{i}$ in the local policy or Q-value function.

\subsubsection{Action Space $\mathcal{A}$}
\label{subsec3}
The control commands of the non-holonomic robot consist of its linear velocity $v^{i}$ and angular velocity $w_{i}$. The both are subject to  the constraints $|v^{i}| \le v_{\max}$ and $|w^{i}| \le w_{\max}$, where $v_{\max}$ and $w_{\max}$ are the maximum line and angular velocities, respectively. Thus, the robot's action can be represented by $a^{i} = [v^{i}, w^{i}] \in \mathcal{A}_{i}$. The joint action $a \in \mathcal{A} $ is defined as $a = [a^{0}, a^{1}, ..., a^{n}]$.

\subsubsection{Extrinsic Reward Function}
\label{subsec6}
The leader’s primary objective is to navigate towards the goal position while avoiding obstacles such as other robots and pedestrians. Then, its extrinsic reward function is given by
\begin{equation}\label{lr}
    \acute{r}_{t}^{0}= \left\{ 
    \begin{aligned}
        & -0.25,   & {\rm if} \ d^{0}_{\min} < 0 \\ 
        & 0.5d^{0}_{\min}-0.1,   & {\rm else \ if} \ d^{0}_{\min} < 0.2 \\ 
        & 100,   & {\rm else \ if} \ p^{0} = \bar{p} \\ 
        & 0,   & {\rm otherwise} \\ 
    \end{aligned}
    \right.
\end{equation}
where $d^{0}_{\min}$ denotes the minimum separation distance between the leader and other agents within a time interval $\Delta t$.

In contrast, the followers' task focuses on maintaining formation and avoiding collisions. Hence, the extrinsic reward function for the follower $i$ is designed by
\begin{equation}\label{fr}
    \acute{r}_{t}^{i}= \left\{
    \begin{aligned}
        & -0.25,   & {\rm if} \ d^{i}_{\min} < 0 \\ 
        & 0.5d^{i}_{\min}-0.1,   & {\rm else \ if} \ d^{i}_{\min} < 0.2 \\ 
        & 1,   & {\rm else \ if} \ 0 \le e_{t}^{i} < 0.2 \\ 
        & -\tanh (7.5e_{t}^{i}-3),   & {\rm else \ if} \ 0.2 \le e_{t}^{i} < 1 \\ 
        & -1,   & {\rm else \ if} \ 1 \le e_{t}^{i} < 2 \\ 
        & -2,   & {\rm otherwise} \\ 
    \end{aligned}
    \right.
\end{equation}
where $d^{i}_{\min}$ denotes the minimum separation distance between follower $i$ and other agents within a duration of $\Delta t$, $e_{t}^{i}=\Vert p^{i} - p^{0} - f^{i} \Vert$ is the formation error with position offset $f^{i}$ of robot $i$ relative to the leader.

\begin{figure}[htbp]
    \centering
    \includegraphics[scale=0.50, trim={13mm 130mm 0mm 15mm}]{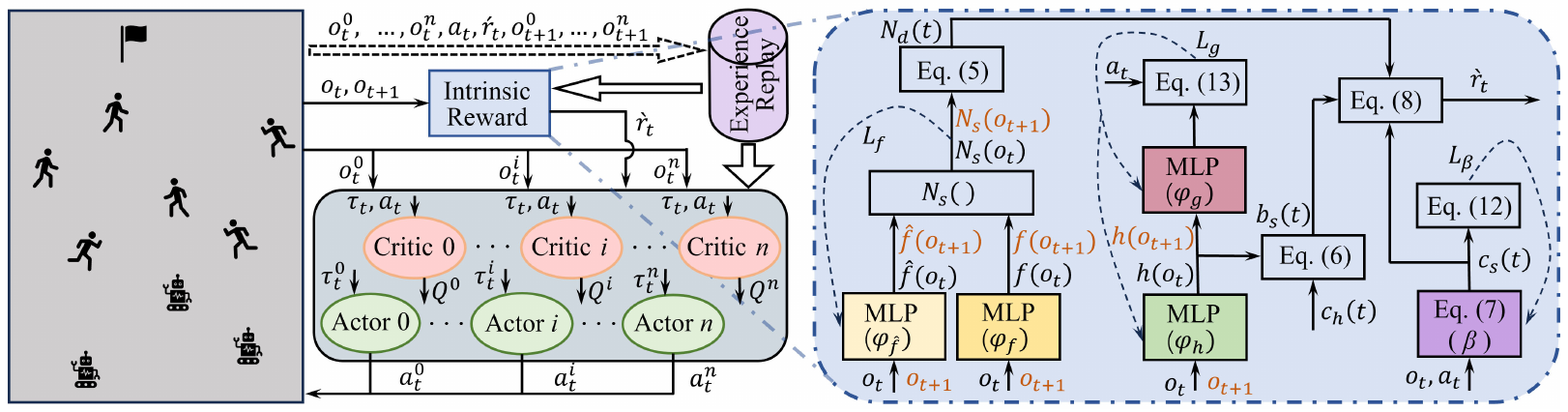}
    \caption{The overall framework of our CEMRRL algorithm.}
    \label{framework}
\end{figure}

In general, maximizing the cumulative discounted extrinsic rewards about (\ref{lr}) and (\ref{fr}) solves the optimization problem (\ref{optim}) by using MARL. Although the extrinsic reward functions (\ref{lr}) and (\ref{fr}) are elaborately designed, the positive reward signals received are very sparse because of the non-stationary of the multi-robot RL and non-cooperative pedestrians' behavior. This can lead to ill-coordinated exploration in the joint-state space $\mathcal{S}$ inevitably, further suffering in the relative overgeneralization problem. In order to drive efficient coordinated exploration for better human-robot interaction efficiency, this paper attempts to propose a CEMRRL algorithm, whose concrete overall framework is illustrated in Figure~\ref{framework}. This algorithm introduces an auxiliary reward, referred to as the intrinsic reward, which augments the previously defined extrinsic rewards (\ref{lr}) and (\ref{fr}), ultimately discovering new optimal solutions for our task. 

\subsection{Self-Learning Intrinsic Reward Mechanism}

This paper designs an intrinsic reward to encourage robots to engage in more coordinated exploration by combining the episodic exploration bonus and the self-adjusting joint policy entropy for visiting novel joint states. Firstly, inspired by the random network distillation~\cite{YuriB2019}, a target embedding function $f(o_{t})$ and its prediction function $\hat{f}(o_{t})$, where $o_{t}=[o_{t}^{0}, o_{t}^{1}, \dots, o_{t}^{n}]$ denotes a joint observation, are introduced.

Building on the concept of the novelty difference~\cite{ZhangT2021}, one defines a life-long novelty differential function:
\begin{align} \label{Novelty}
    N_{d}(t)=\max [N_{s}(o_{t+1})-\alpha N_{s}(o_{t}),\ 0],
\end{align}
where $\alpha$ is a scaling factor and $N_{s}(o_{t})=\Vert \hat{f}(o_{t}) - f(o_{t}) \Vert$.

To remedy dependence of the count-based episodic term, an episodic exploration bonus is computed by
\begin{align}\label{bonus}
    b_{s}(t)=h^{T}(o_{t}) c_{h}(t) h(o_{t}),
\end{align}
where $h(\cdot)$ is a state embedding function, $c_{h}(t)=(\sum_{k=1}^{t-1}h(o_{k})h^{T}(o_{k})) + \lambda I)^{-1}$, and $\lambda$ is a scalar coefficient.

In addition, in order to encourage policy exploration, we also define a joint policy entropy as 
\begin{align} \label{entropy}
    c_{s}(t)=-\beta \log(\pi(a_{t}|o_{t})),
\end{align}
where $\beta$ an adaptively adjusted temperature parameter.

Intuitively, the synergistic integration of the life-long novelty function $N_{d}$, episodic exploration bonus $b_{s}$, and joint policy entropy $c_{s}$ enables the navigation robots to achieve comprehensive exploration through discovery of unseen state space, generation of diverse trajectories, and maintenance of broad action distribution. The former two terms primarily promote diversity within the state space, while the latter term focuses on encouraging diversity in the action space.  To enhance coordinated exploration efficiency,  we formulate an intrinsic reward as follows
\begin{align} \label{inr}
    \grave{r}_{t} =  \sqrt{2b_{s}(t)}N_{d}(t) + c_{s}(o_{t}).
\end{align}

\begin{remark}
As illustrated in Equation (\ref{inr}), the first term integrates a novelty differential function (\ref{Novelty}) into the episodic exploration bonus. This integration not only motivates the robots to explore novel joint states rarely seen in their collective experience, but also promotes the generation of heterogeneous cooperative joint trajectories. Moreover, its multiplicative combination is crucial for achieving efficient synergy, mitigating exploration bias that may arise from their conflicting signals.  The subsequent term ensures the preservation of diversity in the joint action distributions, preventing the joint policy from prematurely converging to repetitive or myopic behaviors. Its adaptively adjusted temperature parameter $\beta$ also contributes to mitigating exploration bias. By synergistically combining the episodic exploration bonus with a self-regulating entropy mechanism for the joint policy, the intrinsic reward mechanism facilitates the emergence of novel joint states, heterogeneous cooperative joint trajectories, and diverse joint action distributions. This allows the robots to collect more cooperative behavioral experience across a broad spectrum of human-robot interactions. Consequently, the collected cooperative behavioral experience mitigates convergence risks associated with the relative overgeneralization suffering from the sparse extrinsic reward and ill-coordinated exploration, thereby reducing overly conservative navigation behaviors.\end{remark}

In addition to the extrinsic rewards about (\ref{lr}) and (\ref{fr}), maximizing the cumulative discounted intrinsic reward (\ref{inr}) allows robots to engage in more coordinated exploration behaviors. Then, the optimization problem (\ref{optim}) can be converted into maximizing the cumulative discounted intrinsic reward (\ref{inr}) in conjunction with the cumulative discounted extrinsic rewards (\ref{lr}) and (\ref{fr}). Hence, the overall objective is to learn a joint policy by maximizing the following objective
\begin{align} \label{optpi}
    \pi^{*} = \arg \max_{\pi}  \sum_{t} \gamma^{t} (\grave{r}_{t} + \sum_{i=0}^{n} \acute{r}^{i}_{t}),
\end{align}
where $\gamma$ is a discount factor, the joint policy  $\pi : \mathcal{S} \to P(\mathcal{A})$ is the collection of all individual policies, i.e., $\pi=\{\pi^{0}, \pi^{1},..., \pi^{n}\}$, and $\pi^{*}=\{\pi^{0*}, \pi^{1*},..., \pi^{n*}\}$ is the optimal joint policy with which $\pi^{i*}$ denotes the optimal policy of Robot $i$. Additionally, to accurately measure the coordinated exploration capacity, it is necessary to learn the feature representations of the intrinsic reward.

\subsection{MARL Framework}
\label{subsec5}

To derive the optimal joint policy $\pi^{*}$, this section integrates the self-learning intrinsic reward within the CTDE framework, facilitating updates for both the multi-agent actor-critic model and the intrinsic reward. Leveraging a two-time-scale update rule and a dual sampling mode learns the optimal multi-robot navigation policy and the feature representations of the intrinsic reward.

Our goal can be reformulated to maximize the expected discounted returns
\begin{align} \label{ret}
    J^{i}(\pi^{i})=\mathbb{E}_{a_{t}^{0}\sim \pi^{0},\ldots, a_{t}^{n}\sim \pi^{n}, s_{t}\sim\mathcal{P}} \sum_{t} \gamma^{t \cdot \bar{v}^{i}} (\grave{r}_{t} + \acute{r}^{i}_{t}),
\end{align}
where $\gamma$ is a discount factor and the joint policy  $\pi : \mathcal{S} \to P(\mathcal{A})$ is the collection of all individual policies, i.e., $\pi=\{\pi^{0}, \pi^{1},..., \pi^{n}\}$.

The policy $\pi^{i}$ is parameterized by $\theta_{i}$, for $i=0, 1, \ldots, n$. Applying the policy gradient techniques, we can write the gradient of the expected discounted return (\ref{ret}) as
\begin{align} \label{gra}
    \nabla_{\theta_{i}} J_{a}^{i}=\mathbb{E}_{\{a_{t}, \tau_{t}\}\sim\mathcal{D}} (\nabla_{\theta_{i}} \log \pi^{i}(a_{t}^{i}|\tau_{t}^{i}; \theta_{i})Q^{i}(\tau_{t},a_{t})),
\end{align}
where $\mathcal{D}$ is an experience replay buffer,
$Q^{i}(\tau_{t},a_{t})$ is a centralized action-value function, $\tau_{t}$ denotes the joint action-observation history or the joint action-observation trajectory. Similarly, $Q^{i}(\tau_{t},a_{t})$ is parameterized by $\phi_{i}$. Then, $Q^{i}(\tau_{t},a_{t}; \phi_{i})$ is updated by
\begin{align}\label{lqi}
    L_{qi}= & \mathbb{E}_{\{a_{t}, \tau_{t}, \grave{r}_{t}, \acute{r}^{i}_{t}, \tau_{t+1}\}\sim\mathcal{D}} (Q^{i}(\tau_{t},a_{t}; \phi_{i}) \nonumber \\
    & - \grave{r}_{t} - \acute{r}^{i}_{t}- \gamma^{t \cdot \bar{v}^{i}} Q^{i}(\tau_{t+1},a_{t+1}; \phi_{i}^{\prime})),
\end{align}
where $\phi_{i}^{\prime}$ is the target parameter.

\begin{remark} Since the hidden states of other robots and pedestrians (for instance, their preferred speeds and heading angles) are not directly observable in practical settings, decision-making must rely solely on locally acquired observational data in the social formation navigation environment. Meanwhile, the continuous interactions among multiple robots and pedestrians dynamically reshape the environment, leading to nonstationarity. In fact,  the actor-critic parameters are updated by sampling both local and joint action-observation trajectories, namely, $\tau_{t}^{i}$ and $\tau_{t}$ as described in~\cite{FoersterJ2018}. In other words, these trajectory sequences in each episode are drawn from the experience replay buffer $\mathcal{D}$ to train the actor-critic parameters $\theta_{i}$ and $\phi_{i}$. This helps the centralized critic and decentralized actors learn to implicitly infer hidden states and respond to the collective environment dynamics by capturing the long-term dependencies and cooperative interactions among multiple robots and pedestrians, mitigating the impact of the non-stationarity. Consequently, the sequential action-observation trajectory information is critical for the centralized training process under the CTDE framework, as it preserves the context of human–robot interactions over time, thereby informing more cohesive policy updates.\end{remark}

To match an entropy target in expectation automatically, we adjust the temperature by the following loss function
\begin{align} \label{beta}
    L_{\beta}=\mathbb{E}_{a_{t}\sim\pi} (c_{s}(t) - \beta \bar{\mathcal{H}}),
\end{align}
where $\bar{\mathcal{H}}$ represents our target joint policy entropy.

As for the intrinsic reward, $f(o_{t})$, $\hat{f}(o_{t})$, $h(o_{t})$, and the inverse dynamics $g(h(o_{t}), h(o_{t+1}))$ can be parameterized as  $f(o_{t}; \varphi_{f})$, $\hat{f}(o_{t}; \varphi_{\hat{f}})$, $h(o_{t}; \varphi_{h})$, and $g(h(o_{t}), h(o_{t+1}); \varphi_{g})$ with the parameters $\varphi_{f}$,  $\varphi_{\hat{f}}$,  $\varphi_{h}$,  and $\varphi_{g}$, respectively. The target embedding network $f(o_{t}; \varphi_{f})$ is randomly initialized and kept fixed throughout the training process. The architecture of the trainable predictor network $\hat{f}(o_{t}; \varphi_{\hat{f}})$ is designed to mirror that of the target embedding network $f(o_{t}; \varphi_{f})$. Our objective is to identify joint observations that are less predictable by the trained predictor network $\hat{f}(o_{t}; \varphi_{\hat{f}})$, considering these observations as more novel. The corresponding component within the self-learning intrinsic reward mechanism motivates the navigation robots to explore states characterized by low visit frequency or high unpredictability in the joint observation space. For this purpose, the parameters of $\hat{f}(o_{t}; \varphi_{\hat{f}})$ are updated by minimizing the loss function $L_{f} = \mathbb{E}_{o_{t}\sim\mathcal{D}} (N_{s})$.

Unlike the target embedding network $f(o_{t}; \varphi_{f})$, the state embedding network $h(o_{t}; \varphi_{h})$ is trained using an inverse dynamics model. This training aims at learning an embedding space where the joint actions are predictive of the change in the embedding of the joint observation. The corresponding component within the self-learning intrinsic reward mechanism promotes the generation of heterogeneous cooperative joint trajectories. Accordingly, measuring the discrepancy between the prediction of the inverse dynamics and the actual action leads to the following loss function
\begin{align} \label{lg}
    L_{g} = \mathbb{E}_{\{o_{t},a_{t},o_{t+1}\}\sim\mathcal{D}} (g(h(o_{t}), h(o_{t+1}); \varphi_{h}, \varphi_{g})- a_{t})^{2}.
\end{align}

\begin{remark}
In essence, the novelty measurement of the self-learning intrinsic reward mechanism relies on  the most recent and relevant transitions to capture environmental changes. Accordingly, this paper updates the intrinsic reward parameters $\varphi_{h}$, $\varphi_{g}$, $\beta$, and $\varphi_{\hat{f}}$ by sampling transition tuples from the same experience replay buffer $\mathcal{D}$, providing a more fine-grained and instantaneous perspective on the human-robot interactions and facilitating prompt adjustments to the intrinsic reward parameters. This allows our self-learning intrinsic reward mechanism to  respond rapidly and adapt based on immediate feedback, thereby reducing policy conservatism.\end{remark}

When simultaneously updating the actor-critic parameters and the intrinsic reward parameters, actor-critic updates occur before the intrinsic reward has sufficiently learned from samples. Subsequently, changes in the actor-critic component alter the data distribution for the intrinsic reward. This means that, the formation robots tend to exploit superficially good joint policies without sufficient coordinated exploration, thereby leading to the relative overgeneralization. 

To alleviate this issue, we leverage a two-time-scale update rule by decoupling interdependent updates through a dual-sampling mode. Specifically, the intrinsic reward parameters are updated by sampling transition tuples under the centralized training in a fast time scale, utilizing a time-varying step size $\kappa_{f}$ with respect to episodes via a polynomial decay rule~\cite{GuptaH2019}. By contrast, updating the actor-critic parameters $\theta_{i}$ and $\phi_{i}$ samples both local and joint action-observation trajectories under the CTDE manner in a slow time scale, with a step size $\kappa_{s}$ much smaller than $\kappa_{f}$. From the viewpoint of the faster time-scale update for the intrinsic reward parameters, the slower update for the actor-critic parameters appears to be quasi-static. In the quasi-stationary situation, updating intrinsic reward parameters allows real‐time adaptation to environmental changes, encouraging sufficient coordinated exploration in the social scenarios. This also provides diverse cooperative action-observation trajectories for updating the actor-critic parameters. Nevertheless, from the viewpoint of the slower time-scale update for the actor-critic parameters, the faster recursion appears to have better convergence. This indicates that these actor-critic parameters evolve gradually through sufficient coordinated exploration experience, thereby obviating the relative overgeneralization issue. 

The training procedure of our CEMRRL algorithm is detailed in Algorithm \ref{alg2}. The intrinsic reward parameters are adjusted at every timestep, whereas the actor-critic parameters are updated at every episode. This configuration amplifies the obvious divergence between the two time scales inherent in the learning process, thereby facilitating more effective decoupling of the two interdependent updates.

\begin{algorithm}[t]
    \caption{CEMRRL}
    \label{alg2}
    \begin{algorithmic}[1]
        \State \textbf{Input} agent number $n$, maximum timestep $T_{\max}$, episode number $M$, mini-batch size
        \State \textbf{Initialize} $\theta_{i}$, $\phi_{i}$, fixed target embedding parameter $\varphi_{f}$, $\varphi_{\hat{f}}$, $\varphi_{h}$, $\varphi_{g}$, replay buffer $\mathcal{D}$, step size sequences $\{\kappa_{f} \}$ and $\{\kappa_{s} \}$

        \For{$episode = 1\ to \ M$}
            \State Initialize trajectory $T_{\tau}=[\,]$ \Comment{empty list}
            \For{$t = 1\ to \ T$}
            \State Select action $a^{i}\sim\pi(a^{i}|o^{i}; {\theta_{i}})$ for each robot $i$
            \State Execute actions $a=[a_{t}^{0}, \ldots, a_{t}^{n}]$ and obtain extrinsic reward $\acute{r}_{t}=[\acute{r}^{0}_{t},\ldots,\acute{r}^{n}_{t}]$, new observation $o_{t+1}=(o^{0}_{t+1},\, ...,\,o^{n}_{t+1})$ and new joint state $s_{t+1}$
            \State Compute intrinsic reward $\grave{r}_{t}$ via (\ref{Novelty})-(\ref{inr})
            \State $T_{\tau} +=[o_{t}^{0},\ldots,o_{t}^{n},a_{t}, \grave{r}_{t},\acute{r}_{t},o_{t+1}^{0},\ldots,o_{t+1}^{n}]$
            \State Sample a random mini-batch transition tuples from $\mathcal{D}$
            \State Update $\varphi_{h}$ and $\varphi_{g}$ by minimizing (\ref{lg}) via $\kappa_{f}$
            \State Update $\varphi_{\hat{f}}$ by minimizing $L_{f}$ via $\kappa_{f}$
            \State Update $\beta$ by minimizing (\ref{beta}) via $\kappa_{f}$
            \EndFor
        \State Store trajectory $T_{\tau}$ in replay buffer $\mathcal{D}$
        \State Sample a random mini-batch action-observation trajectories from $\mathcal{D}$
         \State Update $\phi_{i}$ by minimizing (\ref{lqi}) with the sampled data for each robot $i$ via $\kappa_{s}$
        \State Update $\theta_{i}$ according to the gradient (\ref{gra}) for each robot $i$ via $\kappa_{s}$
        \EndFor
    \end{algorithmic} 
\end{algorithm}

\begin{remark}
The core of our CEMRRL algorithm lies in the introduction of intrinsic motivation exploration guided via a self-learning intrinsic reward mechanism. Specifically, the integration of a life-long novelty function, an episodic exploration bonus, and joint policy entropy promotes the emergence of novel joint states, heterogeneous cooperative joint trajectories, and diverse joint action distributions. This integration effectively mitigates ineffective exploration loops. Furthermore, its incorporation into the CTDE framework consequently yields an improvement in coordinated exploration efficiency through the shared joint state-action information. The coordinated exploration helps the robots escape local traps or loops that would not be broken with uncoordinated individual exploration, thereby promoting adaptation to highly random or adversarial pedestrian motion patterns. Additionally, our CEMRRL algorithm employs a dual-sampling mode and a two-time-scale update rule within the CTDE framework, contributing to implementation of this mechanism in the multi-robot social formation navigation.
\end{remark}

\section{Experiment}
% \label{sec4}
To verify the performance of our algorithm, we have conducted comparison experiments with the state-of-the-art (SOTA) methods.

\subsection{Simulation Setup}
\label{subsec2}
A simulation environment is created in Python for the multi-robot social formation navigation to evaluate the performance of our algorithm. A team of robots with one leader and two followers are configured for the prescribed triangle formation maintenance in the social navigation scenario, which is set up with 5 pedestrians. The initial and goal positions of the robots are determined by an initial distribution. The pedestrians are randomly generated along a circle with radius of 5 m and controlled by ORCA. Additionally, a small amount of Gaussian noise $\mathcal{N}(0, \sigma^{2})$ with $\sigma = 0.05$ is also added to observations and actions for the robustness of our algorithm.

\subsubsection{Baselines}
%% Inline mathematics is tagged between $ symbols.
To rigorously assess the performance of our algorithm, some SOTA methods are selected as baseline methods for comparison. To ensure a fair comparison, some modifications are made for these methods. For instance, MR-Att-RL~\cite{SongC2024} is a representative $\epsilon$-greedy version using self-attention mechanism to extract environmental features. To verify the performance of our coordinated exploration, our CEMRRL combines with the self-attention mechanism, yielding the variant CEMRRL-Att. In addition, MR-SAC~\cite{HeZ2023} is a recent advancement to achieving the SOTA policy exploration in the multi-robot social navigation. The comparison between CEMRRL and MR-SAC is conducted to verify their exploration efficiency.

\subsubsection{Metrics}

To quantify the performance of these different algorithms, the typical representative metrics including success rate, collision rate, and navigation time, are selected for comparison analysis. Additionally, average formation error (AFE), which is the mean of the formation errors during the navigation time, is used to measure the formation maintenance capacity.

\subsubsection{Training Details}
All the aforementioned algorithms are trained using the same set of environmental hyperparameters. The embedding networks $f(o_{t};\phi_{f})$, $\hat{f}(o_{t};\phi_{\hat{f}})$, and $h(o_{t};\phi_{h})$ are fully connected networks with the unit dimension (42, 128, 16), (42, 128, 16), and  (42, 128, 128, 128, 16). The inverse network is also a fully connected network with the unit dimension (32, 128, 128, 128, 6). Meanwhile, they are also equipped with the Layernorm layer and the LeakyReLU activation function. 

In the contrast experiments, both CEMRRL-Att and MR-Att-RL are set in the equivalent training settings. CEMRRL and MR-SAC are trained in the same RL framework as well as MLP settings but with different reward components.

\begin{figure}[htbp]
	\centerline{\includegraphics[scale=0.35, trim={0mm 15mm 0mm 12mm}]{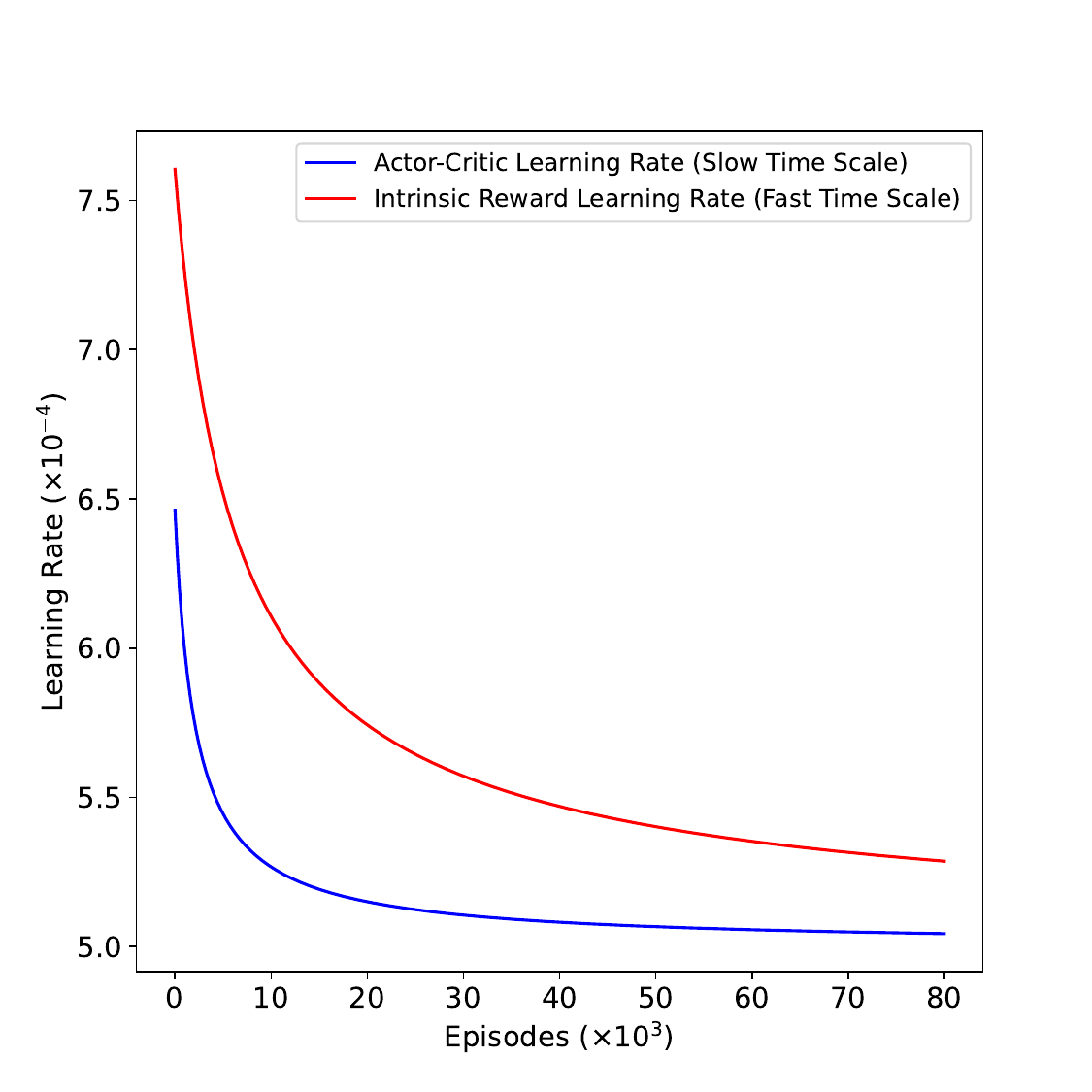}}
	\caption{ Learning rates in the actor-critic learning and 
 the intrinsic reward learning.}
	\label{Learningrates}
\end{figure}

The robots obtain their decision policies by using our proposed algorithm, which is implemented in Pytorch with Adam optimizer. The specific hyperparameter values are summarized in Table~\ref{para}. The selection of these parameters is validated through a combination of empirical knowledge and systematic grid search. Note that, the learning rates of the actor-critic parameters and the intrinsic reward parameters are the sum of their step sizes and the basic learning rate in Table~\ref{para}. This basic learning rate ensures continual updates of these parameters with the minimum training pace. Selection of the step sizes is referred to \cite{GuptaH2019}. The evolution of the learning rates is depicted in Figure \ref{Learningrates}. Combining with the time-step update of the intrinsic reward parameters and the episodic updates of the actor-critic parameters in Algorithm \ref{alg2}, actual updating frequency of the actor-critic learning is much less than that of the intrinsic reward learning. This facilitates more effective decoupling of the two interdependent updates.

\begin{table}[htbp]
\footnotesize
\setlength{\abovecaptionskip}{-0.3cm}
\setlength{\belowcaptionskip}{-0.5cm}
	\begin{center}
		\begin{tabular}{lclc}
			\hline
			Parameter  &  Value  & Parameter  &  Value \\
			\hline
			Discount factor $\gamma$ & 0.99 & Initial temperature parameter  &  0.01 \\
			Batch size  & 256 & Replay buffer capacity $\mathcal{D}$ & 2e5 \\
			Scaling factor $\alpha$ & 0.5   & Maximum time limit & 21 s  \\
			Timestep  &  0.25 s  &  Maximum episode  & 8e4  \\
			Basic learning rate  & 5e-4  & Scalar coefficient $\lambda$ &  0.1 \\
			\hline
		\end{tabular}
	\end{center}
        \caption{Hyperparameter}
        \label{para}
\end{table}

In our CEMRRL algorithm, a scaling factor $\alpha$ is crucial for enabling navigation robots to thoroughly explore novel or uncertain regions. Meanwhile, selection of the scalar coefficient $\lambda$ also influences the generation of heterogeneous cooperative joint trajectories. To evaluate their impact on performance, a one-at-a-time sensitivity analysis was conducted for the hyperparameters $\alpha$ and  $\lambda$. The detailed results are summarized in Table~\ref{sens}. It is evident from the results that, in the social formation navigation experiments, the overall performance is significantly superior when $\alpha = 0.5$ and $\lambda = 0.1$. This selection also meets the parameter balance between novelty differential function and the episodic exploration bonus. Accordingly, this specific parameter configuration was utilized in all subsequent quantitative and qualitative evaluations.

\begin{table*}[htbp]
\footnotesize
\setlength{\abovecaptionskip}{-0.3cm}
\setlength{\belowcaptionskip}{-0.5cm}
	\begin{center}{
    \resizebox{\textwidth}{!}{
    \begin{tabular}{cccccc}
        \hline
        Hyperparameter & Value & Success rate (\%) & Collision rate (\%) & Navigation time (s) & AFE \\
        \hline
        \multirow{5}{*}{$\alpha$} & 0.1 &89.5  &10.5 &9.69 &0.80  \\
        \cline{2-6}  
        \multicolumn{1}{c}{} & \multicolumn{1}{c}{0.3} &91.0 &9.0 &9.66 &0.79  \\
        \cline{2-6}   
        \multicolumn{1}{c}{} & \multicolumn{1}{c}{0.5} &91.4  &8.6 &9.56
               &0.77 \\
        \cline{2-6}  
        \multicolumn{1}{c}{} & \multicolumn{1}{c}{1} &90.1  &9.9 &9.71
               &0.81 \\
        \cline{2-6}  
        \multicolumn{1}{c}{} & \multicolumn{1}{c}{2} &85.3  &14.7 &9.79
               &0.86 \\
        \hline
        \multirow{5}{*}{$\lambda$} & 0.01 &86.3  &13.7 &9.68 &0.84 \\
         \cline{2-6} 
         \multicolumn{1}{c}{} & \multicolumn{1}{c}{0.05} &86.6  &13.4 &9.63 &0.83 \\
         \cline{2-6}   
        \multicolumn{1}{c}{} & \multicolumn{1}{c}{0.1} &91.4  &8.6 &9.56
               &0.77 \\
        \cline{2-6}  
        \multicolumn{1}{c}{} & \multicolumn{1}{c}{0.2} &90.3  &9.8 &9.60
               &0.78 \\ 
        \cline{2-6}  
        \multicolumn{1}{c}{} & \multicolumn{1}{c}{0.5} &88.6  &11.4 &9.61
               &0.80 \\ 
        \hline
    \end{tabular}}}
	\end{center}
    \caption{Sensitivity analysis of parameters $\alpha$ and  $\lambda$}
    \label{sens}
\end{table*}

\subsection{Quantitative Evaluation}

\begin{figure}[htbp]
	\centering
	\subfigure{
	\includegraphics[scale=0.259, trim={15mm 10mm 11mm 10mm}]{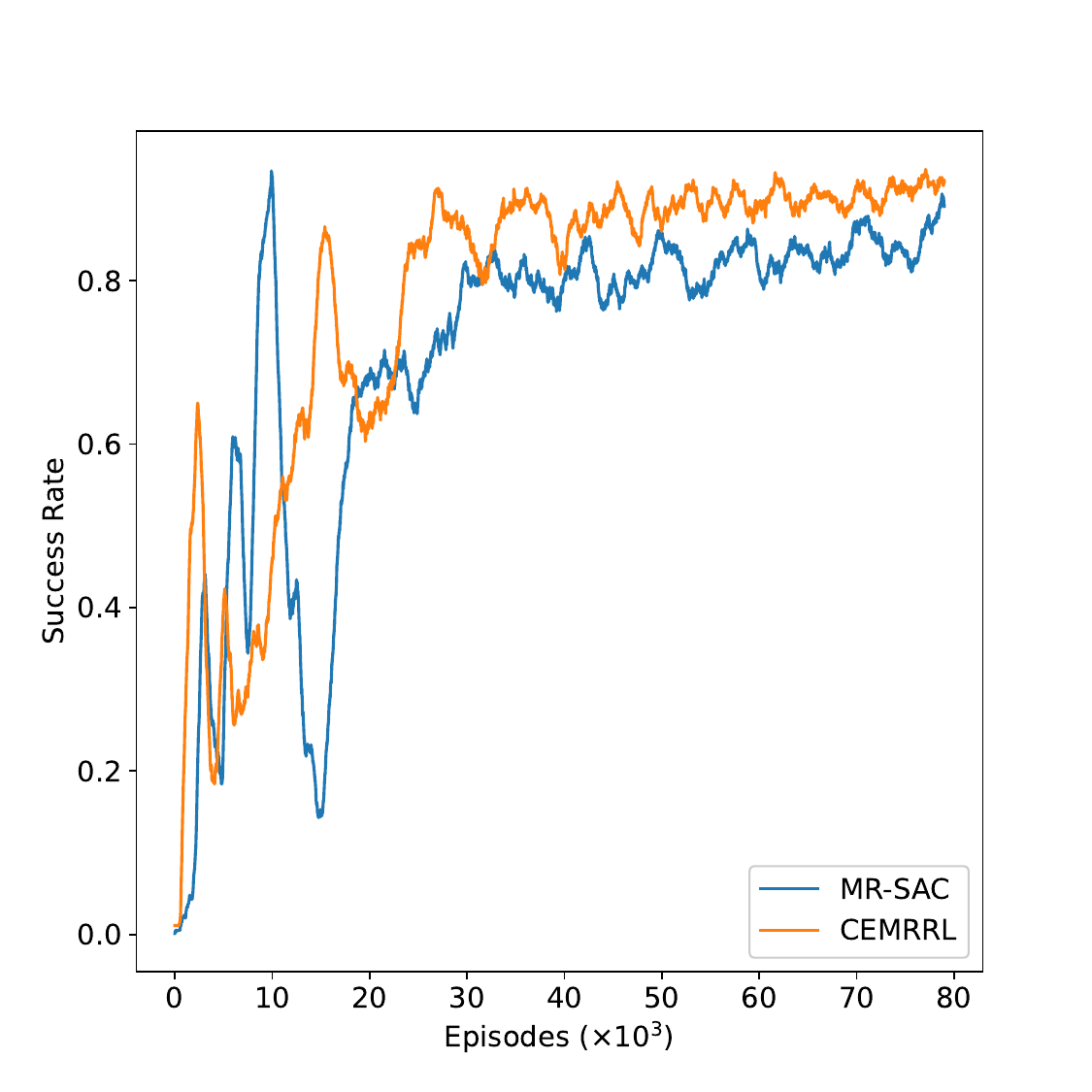}}
        \centering
        \setcounter{subfigure}{0}
	\subfigure[\fontsize{6}{1}\selectfont MR-SAC and CEMRRL]{
	\label{Rewarda}
	\includegraphics[scale=0.259, trim={15mm 10mm 11mm 10mm}]{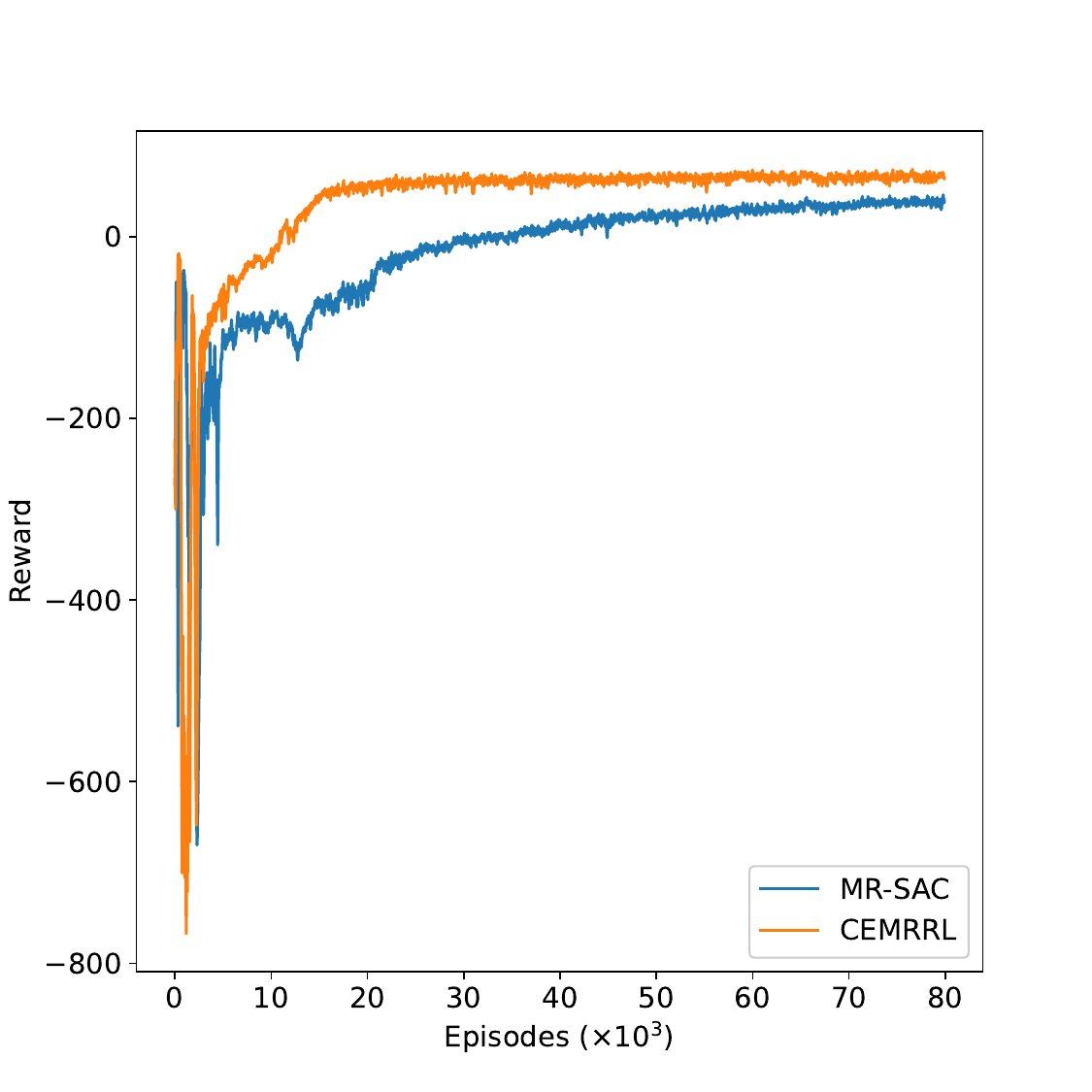}}
	\centering
	\subfigure{
	\includegraphics[scale=0.259, trim={15mm 10mm 11mm 10mm}]{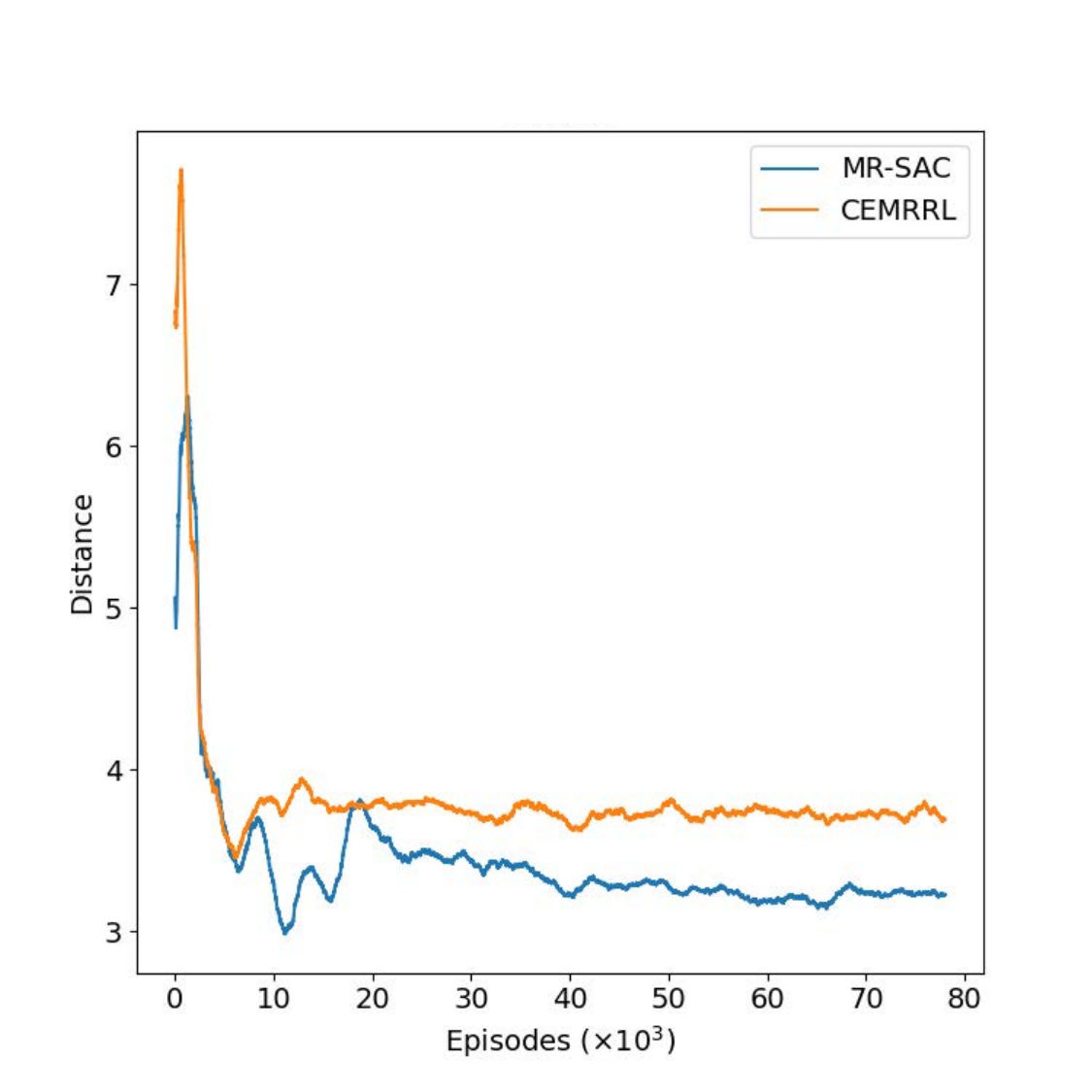}}
    
	\centering
	\subfigure{
	\includegraphics[scale=0.26, trim={15mm 10mm 10mm 10mm}]{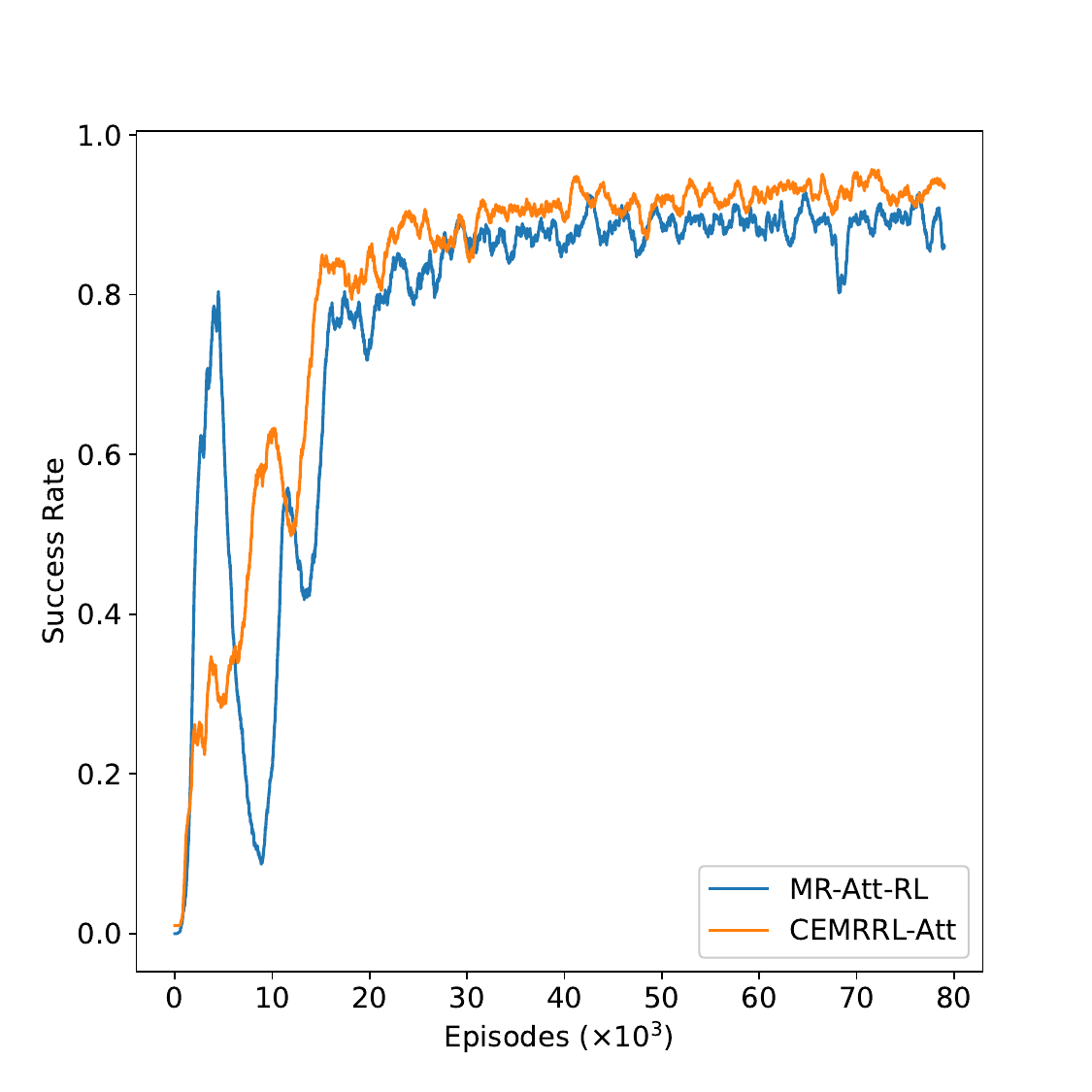}}
        \centering
        \setcounter{subfigure}{1}
	\subfigure[\fontsize{6}{1}\selectfont  {MR-Att-RL and CEMRRL-Att}]{
	\label{Rewardb}
	\includegraphics[scale=0.26, trim={15mm 10mm 10mm 10mm}]{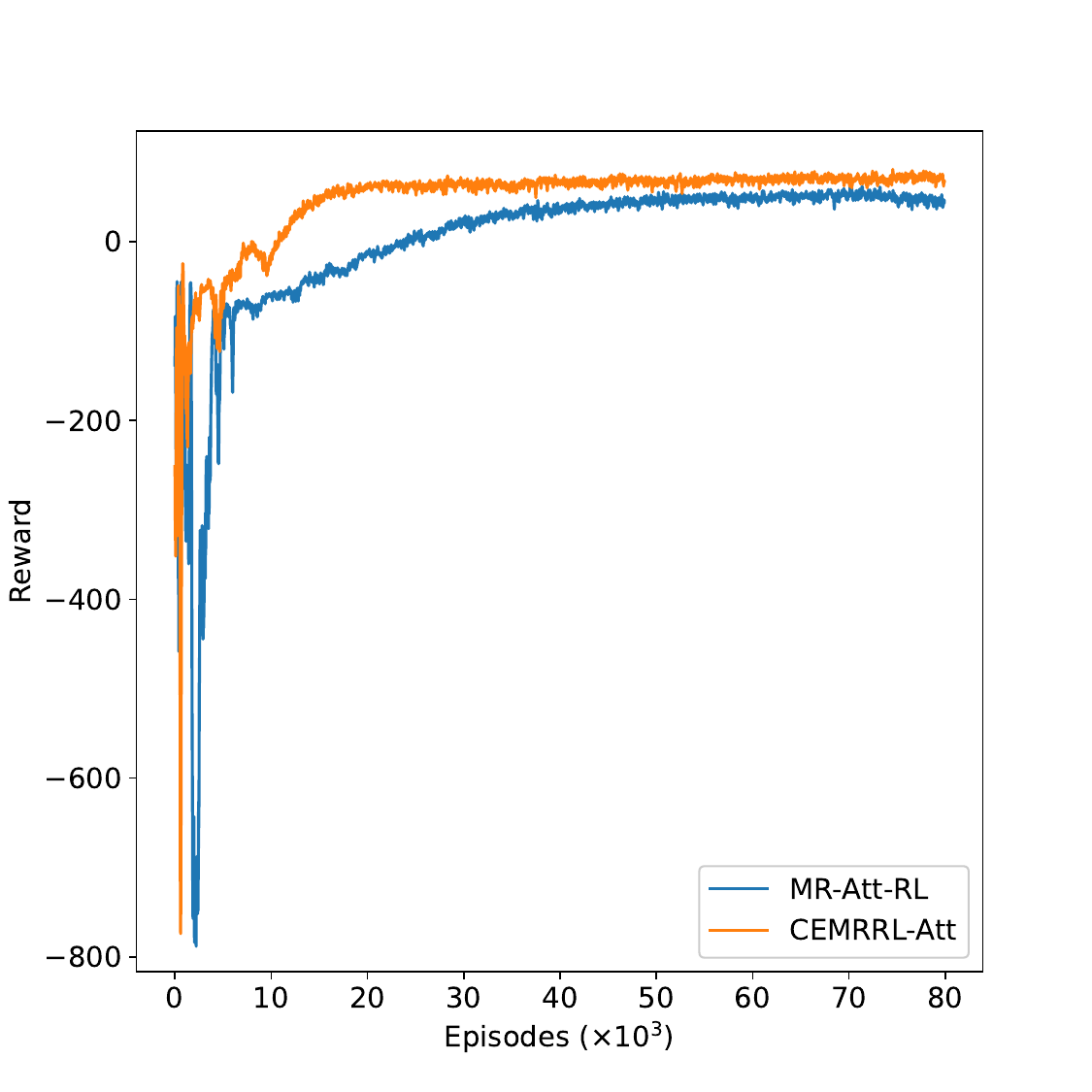}}
        \centering
	\subfigure{
	\includegraphics[scale=0.26, trim={15mm 10mm 10mm 10mm}]{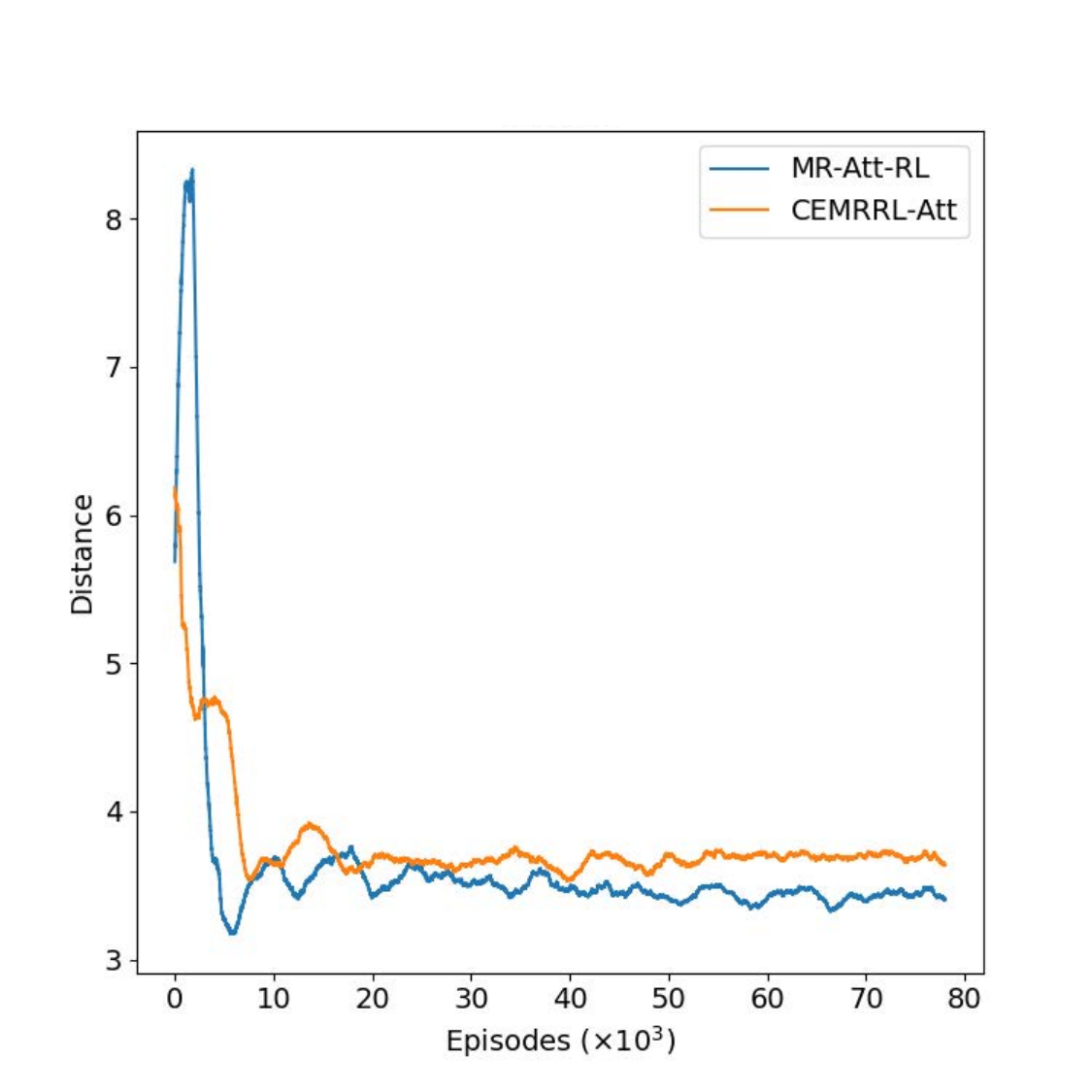}}
	\caption{The training curves throughout the training process. (a) Comparative experiment between CEMRRL and MR-SAC~\cite{HeZ2023} in terms of the success rate, reward, and distance to collision, validating the coordinated exploration capacity facilitated by our self-learning intrinsic reward mechanism and maximum policy entropy. (b) CEMRRL-Att, combination of our CEMRRL algorithm and the attention mechanism, compared with MR-Att-RL~\cite{SongC2024}. }
\label{training}
\end{figure}

Following 80,000 training episodes, the performance curves for all evaluated methods, encompassing success rate, cumulative reward, and minimum distance to collision, are visualized in Figure~\ref{training}. The results show that our algorithm has comparable convergence to the baselines. As illustrated in Figure~\ref{Rewarda},  both methods achieve convergence during training. Notably, our proposed CEMRRL algorithm demonstrates superior convergence of the success rate compared to MR-SAC, despite the latter's well-documented advantage in exploration through maximum policy entropy. From the evolution of the reward curves, we find that our CEMRRL algorithm achieves faster convergence and yields higher rewards compared to MR-SAC. Specifically, our algorithm reaches a convergent state after approximately 20,000 episodes, whereas the baseline method requires approximately 65,000 episodes to converge. Besides, our algorithm maintains a greater minimum distance to collision. Collectively, these results indicate that our CEMRRL algorithm facilitates collecting more efficient coordinated exploration by leveraging human-robot interaction, thereby contributing to faster overall convergence and improved converged performance in the multi-robot social formation navigation. To further demonstrate its extensibility, we investigated the integration of a self-attention mechanism. As shown in Figure~\ref{Rewardb}, this integration yields slight performance improvements for both our CEMRRL algorithm and the baseline. This improvement can be attributed to the fact that the introduction of the attention mechanism facilitates the capture of correlational dependencies within human-robot interaction data. Nonetheless, our CEMRRL algorithm remains superior to the baseline method in overall performance. This suggests that our algorithm possesses significant extensibility for integration with other techniques in the context of social formation navigation.

Based on the trained model, three random testing scenarios (5 pedestrians, 7 pedestrians, and 9 pedestrians) are selected to further validate our algorithm via quantitative analysis. For consistency and fairness, each method is evaluated over 1000 episodes. The test results are summarized in Table~\ref{quant}. We find that the overall performance of all methods consistently tends to be worse as the pedestrian density increases. Intuitively, this is due to the fact that increase of the pedestrian density makes the multi-robot social formation navigation task more difficult in a limited space of environment. It is evident that the evaluation performance of our method only exhibits a slight degradation, which implies that our proposed CEMRRL has advanced scalability and strong robustness in handling large-scale complex various social scenarios. In the same scenario, combining with self-attention mechanism, evaluation performance of our CEMRRL is overall superior to the baseline method by comparison between CEMRRL-Att and MR-Att-RL. This finding indicates the advantages of our derived joint policy in terms of the coordinated exploration capacity from the human-robot interaction, leading to superior social formation navigation performance. It is also evident that the proposed CEMRRL algorithm has also good extensibility to seamlessly integrate with self-attention mechanisms to process input data effectively.

\begin{table*}[htbp]
\footnotesize
\setlength{\abovecaptionskip}{-0.3cm}
\setlength{\belowcaptionskip}{-0.5cm}
	\begin{center}{
    \resizebox{\textwidth}{!}{
    \begin{tabular}{cccccc}
        \hline
        Scenarios & Methods & Success rate (\%) & Collision rate (\%) & Navigation time (s) & AFE \\
        \hline
        \multirow{4}{*}{5} & MR-Att-RL &93.6  &6.4 &9.66
               &0.75 \\
        \cline{2-6}   
        \multicolumn{1}{c}{} & \multicolumn{1}{c}{CEMRRL-Att} &\textbf{95.2}  &\textbf{4.8} &\textbf{9.45} &\textbf{0.61} \\
        \cline{2-6}   
        \multicolumn{1}{c}{} & \multicolumn{1}{c}{MR-SAC} &92.4  &7.6 &9.58
               &0.70 \\
        \cline{2-6}   
        \multicolumn{1}{c}{} & \multicolumn{1}{c}{CEMRRL} &\textbf{94.1}  &\textbf{5.9} &\textbf{9.49}
               &\textbf{0.62} \\               
        \hline
        \multirow{4}{*}{7} & MR-Att-RL &91.2  &8.8 &9.87 &1.03  \\
        \cline{2-6}  
        \multicolumn{1}{c}{} & \multicolumn{1}{c}{CEMRRL-Att} &\textbf{93.0} &\textbf{7.0} &\textbf{9.66} &\textbf{0.91}  \\
        \cline{2-6}   
        \multicolumn{1}{c}{} & \multicolumn{1}{c}{MR-SAC} &90.8 &9.2  &9.74
               &1.01 \\
        \cline{2-6}  
        \multicolumn{1}{c}{} & \multicolumn{1}{c}{CEMRRL} &\textbf{92.3}  &\textbf{7.7} &\textbf{9.68}
               &\textbf{0.96} \\
        \hline
        \multirow{4}{*}{9} & MR-Att-RL &88.8  &11.2 &10.07 &1.40 \\
         \cline{2-6} 
         \multicolumn{1}{c}{} & \multicolumn{1}{c}{CEMRRL-Att} &\textbf{90.6}  &\textbf{9.4} &\textbf{9.87} &\textbf{1.11} \\
         \cline{2-6}   
        \multicolumn{1}{c}{} & \multicolumn{1}{c}{MR-SAC} &88.1  &11.9 &10.10 &1.36 \\
        \cline{2-6}  
        \multicolumn{1}{c}{} & \multicolumn{1}{c}{CEMRRL} &\textbf{89.6}  &\textbf{10.4} &\textbf{9.89}
               &\textbf{1.29} \\ 
        \hline
    \end{tabular}}}
	\end{center}
    \caption{Quantitative evaluation results under different pedestrian participation scenarios (5 pedestrians, 7 pedestrians, and 9 pedestrians)}
    \label{quant}
\end{table*}

\subsection{Qualitative Evaluation}
\begin{figure}[htbp]
\setlength{\abovecaptionskip}{-0.1cm}
\setlength{\belowcaptionskip}{-0.5cm}
	\centering
	\subfigure[MR-Att-RL]{
		\label{Testing2c}
		\includegraphics[scale=0.26, trim={6mm 5mm 6mm 11mm}]{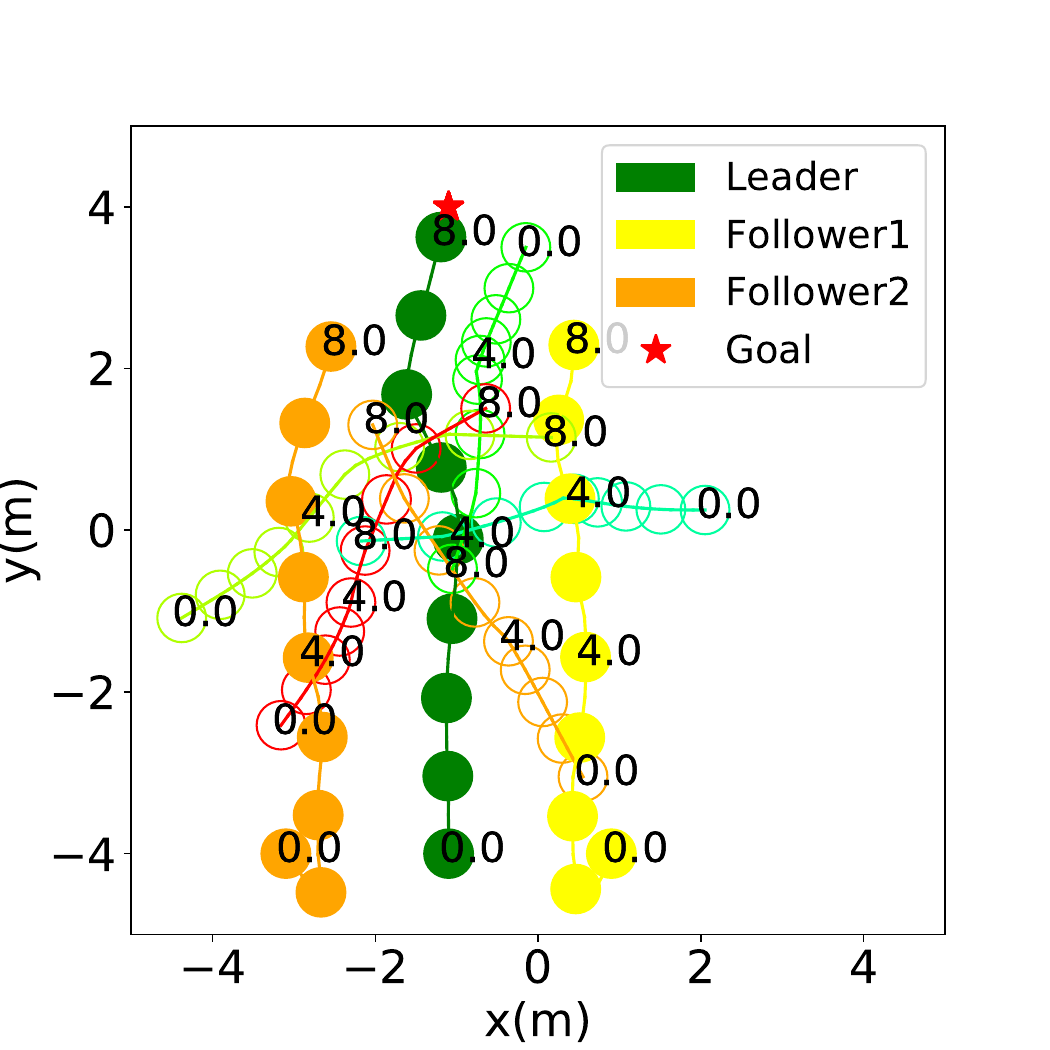}}
	\centering
	\subfigure[CEMRRL-Att]{
		\label{Testing2d}
		\includegraphics[scale=0.26, trim={6mm 5mm 6mm 11mm}]{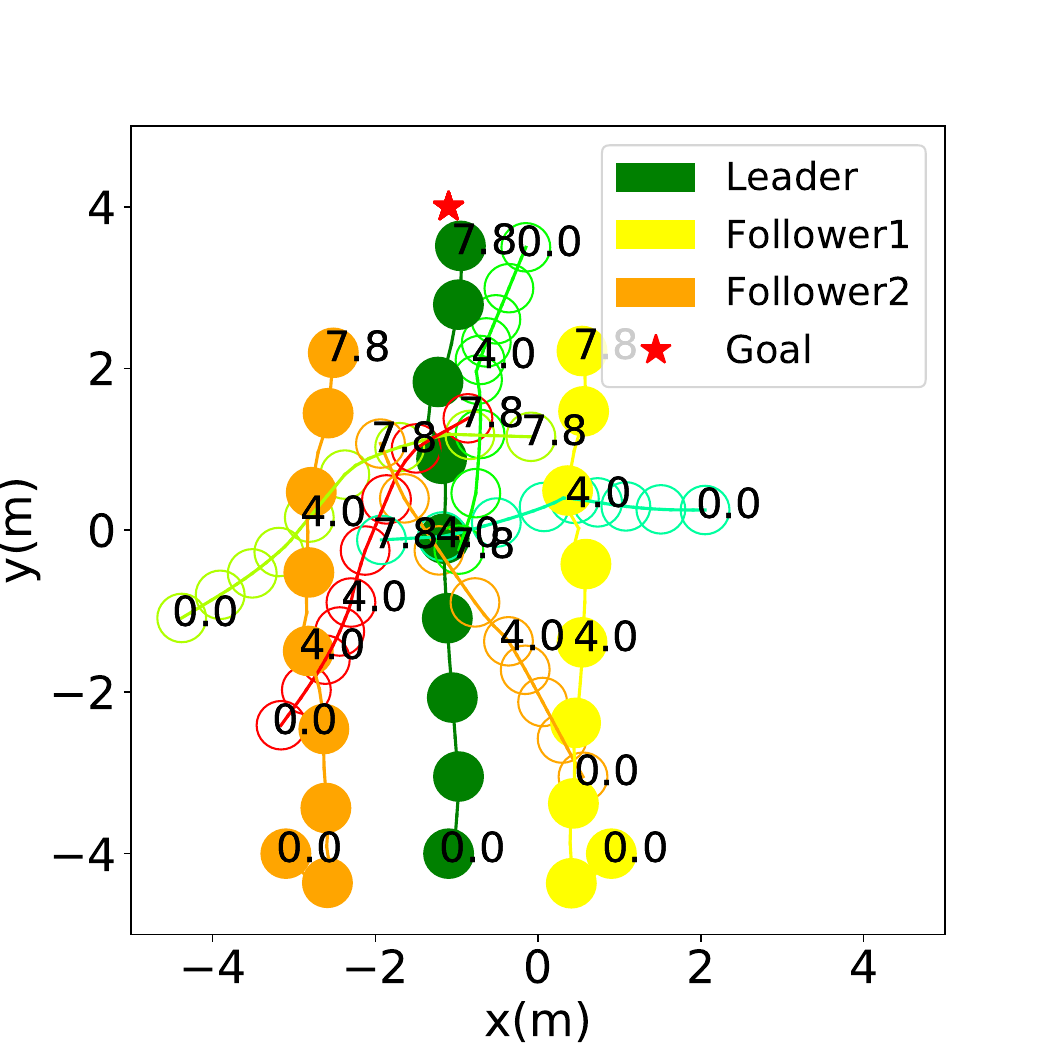}}        
	\centering
	\subfigure[MR-SAC]{
		\label{Testing2e}
		\includegraphics[scale=0.26, trim={6mm 5mm 6mm 11mm}]{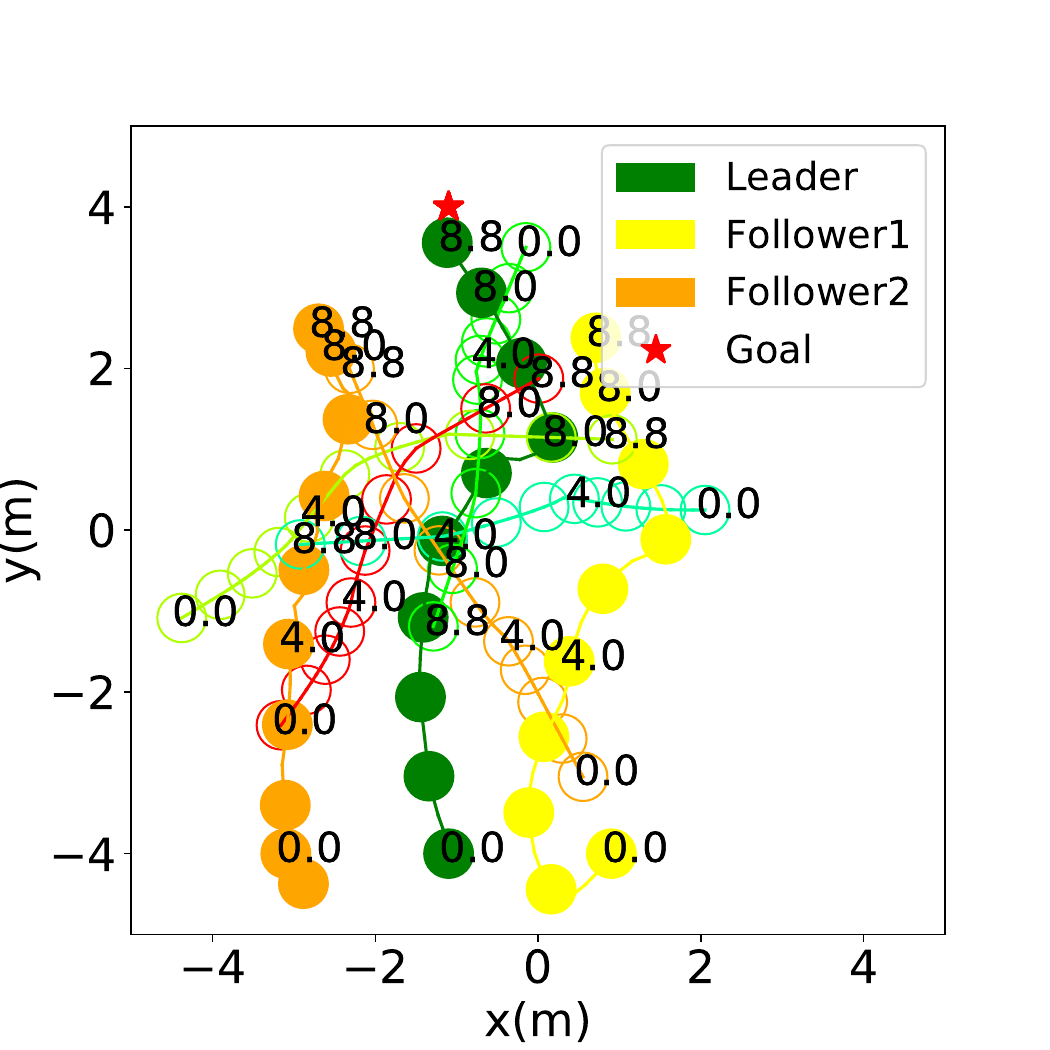}}
        
	\centering
	\subfigure[CEMRRL]{
		\label{Testing2f}
		\includegraphics[scale=0.26, trim={6mm 5mm 6mm 11mm}]{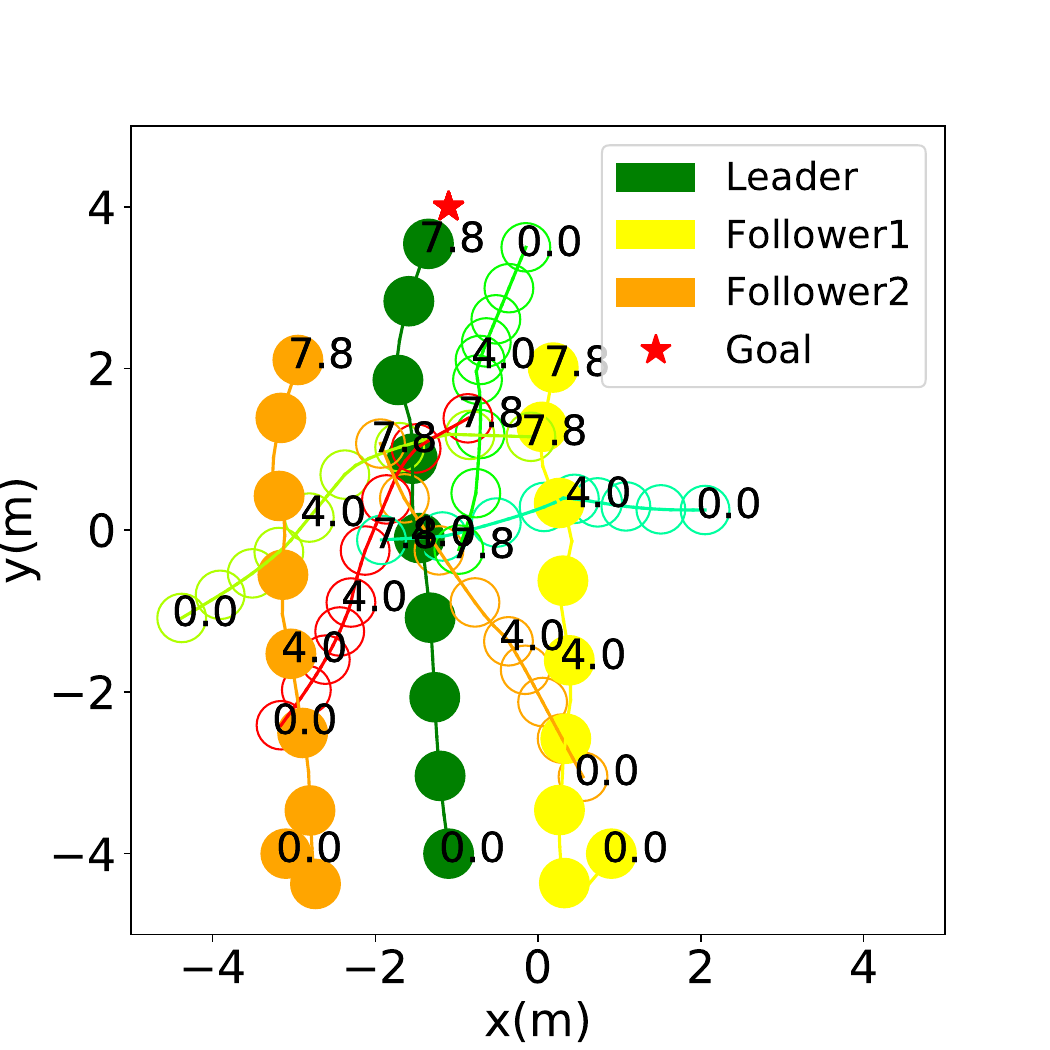}}
        \centering
	\subfigure[CEMRRL for stop]{
		\label{Testing2g}
		\includegraphics[scale=0.26, trim={6mm 5mm 6mm 11mm}]{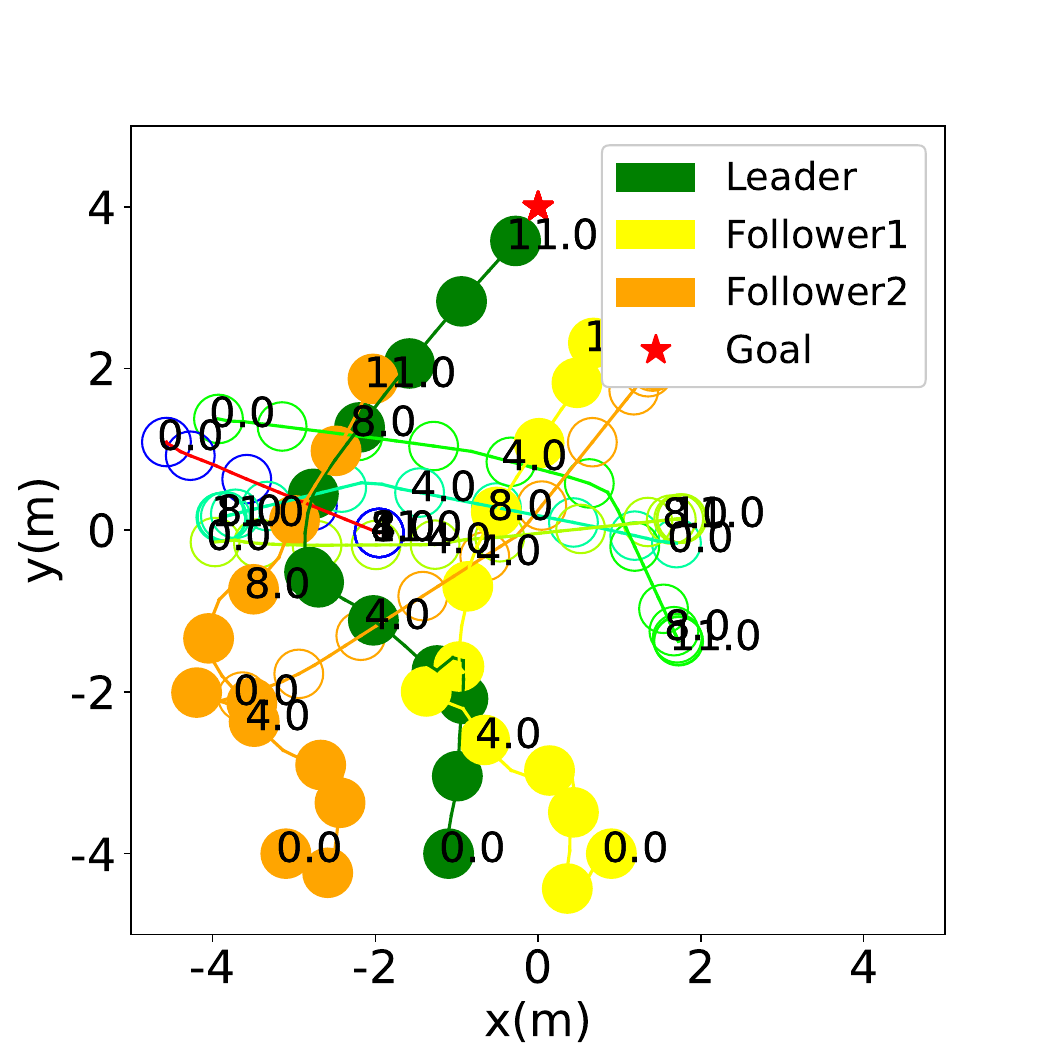}}
	\centering
	\subfigure[CEMRRL for reversal]{
		\label{Testing2h}
		\includegraphics[scale=0.26, trim={6mm 5mm 6mm 11mm}]{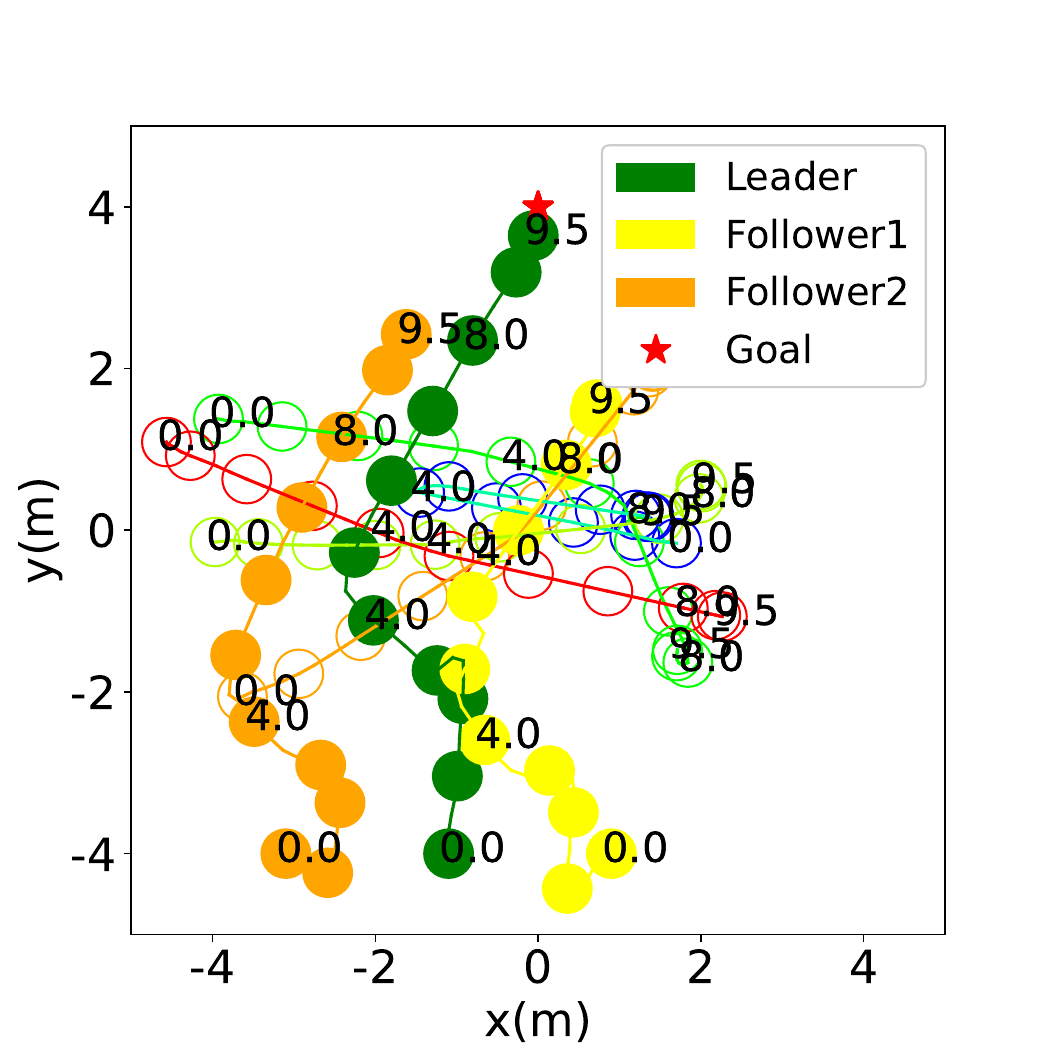}}
	\caption{Multi-robot trajectory comparison of the different methods in identical social formation navigation test scenarios. The green disc denotes the leader robot. The yellow or orange disc denotes the follower robot. The red star stands for the leader goal position. The circles represent pedestrians. The color curves denote their corresponding trajectories. For example, the leader motion trajectory is marked in green.}
	\label{Testing2}
\end{figure}

To further validate our algorithm via qualitative analysis, global trajectories generated by these different methods are used to intuitively visualize the multi-robot behaviors in a shared initial setting and desired formation shape within a randomized multi-robot scenario. The details are depicted in Figure~\ref{Testing2}. As seen in Figure~\ref{Testing2c}, MR-Att-RL motivates the formation robots to avoid the crowds and accomplishes the social formation navigation task. But, due to lack of more efficient exploration, it tends to select a relatively conservative joint policy to maintain an unnecessarily large safety margin. By contrast, thanks to our CEMRRL-Att reducing the conservatism of the joint policy, its multi-robot trajectories shorter in time are acquired in Figure~\ref{Testing2d}.

We further look into the qualitative comparison between our CEMRRL and MR-SAC, whose trajectory evolution is described in Figures~\ref{Testing2e} and \ref{Testing2f}. It can be observed that the formation robots of MR-SAC go around at a low speed when encountering pedestrians. By contrast, our CEMRRL chooses to slow down or bypass pedestrians in advance to avoid the crowds, although this occasionally causes temporary formation deformation. We believe the main reason is that multi-robot policy coupling has an impact on the adjustable temperature in MR-SAC. Accordingly, the ill-coordinated exploration problem arises in the social formation navigation, giving rise to relative overgeneralization issue. Instead, our CEMRRL obviates such an issue by leveraging its inherent merit of intrinsic-motivation coordinated exploration, resulting in the more farsighted navigation policy and cooperation behaviors. 

To ascertain the algorithm's resilience to unprecedented pedestrian behaviors (e.g., sudden stops, rapid direction changes), two dedicated tests are conducted, as depicted in Figures \ref{Testing2g} and \ref{Testing2h}. Observations from these tests reveal that, despite unexpected pedestrian actions such as sudden stops (Figure \ref{Testing2g}) or reversals (Figure \ref{Testing2h}), the robots consistently maintain their formation while successfully circumnavigating all pedestrians. These findings indicate that unprecedented pedestrian behaviors do not detrimentally affect the social formation navigation task. This outcome is primarily attributed to the intrinsic reward mechanism's ability to efficiently learn and represent these unprecedented pedestrian patterns through coordinated exploration, thereby enabling our robots to execute appropriate avoidance maneuvers in subsequent encounters.

\subsection{Ablation Study}
To verify the effectiveness of the three components (novelty function $N_{d}$, episodic bonus $b_s(t)$, and policy entropy  $c_s(t)$) in our self-learning intrinsic reward mechanism, this paper further conducts an ablation study. The corresponding ablation models are denoted as follows: EB-PE  (excluding  $N_{d}$), NF-PE  (excluding  $b_s(t)$), and NF-EB (excluding  $c_s(t)$). These ablation models are independently trained and evaluated under the same experimental conditions as previously described. The results are summarized in Table~\ref{ablation}. A comparison between Table~\ref{quant} and Table~\ref{ablation} reveals that their comprehensive performance on the social formation navigation is slightly superior to that of the baseline presented in Table~\ref{quant}. This finding indicates that the inclusion of each component contributes to marginal improvements in exploration efficiency. Furthermore, the results presented in this table demonstrate that our CEMRRL algorithm outperforms these ablation models across various performance metrics. It can be inferred that our algorithm effectively synthesizes the contributions of these components, leading to significantly more efficient coordinated exploration in the social formation navigation and consequently reducing policy conservatism.

\begin{table*}[htbp]
\footnotesize
\setlength{\abovecaptionskip}{-0.3cm}
\setlength{\belowcaptionskip}{-0.5cm}
\begin{center}{
    \resizebox{\textwidth}{!}{
    \begin{tabular}{cccccc}
        \hline
        Scenarios & Methods & Success rate (\%) & Collision rate (\%) & Navigation time (s) & AFE \\
        \hline
        \multirow{4}{*}{5} & EB-PE  &93.7  &6.3 &9.56 &0.68 \\
        \cline{2-6}   
        \multicolumn{1}{c}{} & NF-PE  &93.9  &6.2 &9.55 &0.66 \\
        \cline{2-6}   
        \multicolumn{1}{c}{} & NF-EB & 93.8  &6.2 &9.52 &0.65 \\
        \cline{2-6}   
        \multicolumn{1}{c}{} & CEMRRL & \textbf{94.1}  &\textbf{5.9} &\textbf{9.49} &\textbf{0.62} \\              
        \hline
        \multirow{4}{*}{7} & EB-PE & 91.4 & 8.7 & 9.72 & 1.01 \\
        \cline{2-6}  
        \multicolumn{1}{c}{} & NF-PE &  91.3  &8.8 &9.74 &0.99  \\
        \cline{2-6}   
        \multicolumn{1}{c}{} & NF-EB & 91.7  &8.3 &9.73 &0.98 \\
        \cline{2-6}  
        \multicolumn{1}{c}{} & CEMRRL &\textbf{92.3}  & \textbf{7.7} & \textbf{9.68} & \textbf{0.96} \\
        \hline
        \multirow{4}{*}{9} & EB-PE &89.0  &11.1 & 10.02 &1.31 \\
         \cline{2-6} 
         \multicolumn{1}{c}{} & NF-PE & 88.9  & 11.0 & 9.98 & 1.33 \\
         \cline{2-6}   
        \multicolumn{1}{c}{} & NF-EB &89.2  &11.0 & 9.96 & 1.34 \\
        \cline{2-6}  
        \multicolumn{1}{c}{} & CEMRRL &\textbf{89.6}  &\textbf{10.4} &\textbf{9.89} &\textbf{1.29} \\ 
        \hline
    \end{tabular}}}
	\end{center}
    \caption{ Ablation experiment results}
    \label{ablation}
\end{table*}

\section{Conclusion}
% \label{sec4}
In this study, we have proposed the CEMRRL algorithm by introducing an intrinsic motivation exploration criteria to overcome the inefficient ill-coordinated exploration issue. Our CEMRRL establishes a self-learning intrinsic reward, allowing our formation robots to learn a lower conservative and safe navigation policy. The experiments show that, our CEMRRL can obtain better social formation navigation performance in terms of the quantitative and qualitative evaluations about the evaluation metrics, including success rate, collision time, and average formation error. In comparison to the existing methods, we have achieved the SOTA results to date.  In our future study, we will focus on analyzing the convergence of our proposed algorithm to better quantify the parameter coupling effect.

\bibliographystyle{elsarticle-num}
\bibliography{reference}

\end{document}